%% file: ms.tex
\documentclass[runningheads]{llncs}
\usepackage{graphicx}
\usepackage{bookmark}
\usepackage{amsmath,amssymb} 
\usepackage{color}
\input{math_commands.tex}

\usepackage{hyperref}
\usepackage{url}
\usepackage{amsfonts}
\usepackage{multirow}
\usepackage{hhline}
\usepackage{booktabs}
\usepackage{makecell}
\usepackage{graphicx}
\usepackage{caption}
\usepackage{subcaption}
\usepackage{amsmath,amstext}
\usepackage{wrapfig}
\usepackage[linesnumbered,ruled,vlined]{algorithm2e}
\DontPrintSemicolon  

\newcommand{\DANAME}{CDA}
\newcommand{\DANAMEFULL}{Class-imbalanced Domain Adaptation} 

\newcommand{\SHIFTNAME}{feature shift}
\newcommand{\YSHIFTNAME}{label shift}
\newcommand{\SHIFTNAMEFULL}{feature shift}
\newcommand{\YSHIFTNAMEFULL}{label shift}
\newcommand{\MODELNAME}{COAL}

\makeatletter
\def\BState{\State\hskip-\ALG@thistlm}
\makeatother
\usepackage{xcolor}

\newcommand{\ssh}[1]{\textcolor{black}{#1}}

\begin{document}

\pagestyle{headings}
\mainmatter
\def\ECCV20SubNumber{2}  
\title{Class-imbalanced Domain Adaptation:\\ An Empirical Odyssey}
\titlerunning{Class-imbalanced Domain Adaptation: An Empirical Odyssey}
\authorrunning{Tan, Peng et al.}
\author{Shuhan Tan$^{1}$ \thanks{Work done while the author was visiting Boston University}, Xingchao Peng$^2$, Kate Saenko$^{2,3}$
}
\institute{$^1$Sun Yat-Sen University, $^2$Boston University, $^3$MIT-IBM Watson AI Lab\\
\texttt{tanshh@mail2.sysu.edu.cn, \{xpeng, saenko\}@bu.edu} \\}

\maketitle

\begin{abstract}
Unsupervised domain adaptation \ssh{is a promising way}
to generalize deep models to novel domains. 
However the current literature assumes that the label distribution is domain-invariant and only aligns the feature distributions or \textit{vice versa}. 
In this work, we explore the more realistic task of \ssh{\textit{\DANAMEFULL}}:
\ssh{How to align feature distributions across domains \textit{while} the label distributions of the two domains are also different?}
\ssh{Taking a practical step towards this problem, we constructed its first benchmark with 22 cross-domain tasks from 6 real-image datasets. We conducted comprehensive experiments on 10 recent domain adaptation methods and find most of them are very fragile in the face of coexisting feature and label distribution shift.}
\ssh{Towards a better solution, we further proposed a \ssh{feature} and \ssh{label} distribution CO-ALignment (COAL) model with a novel combination of existing ideas. COAL is empirically shown to outperform most recent domain adaptation methods on our benchmarks. We believe the provided benchmarks, empirical analysis results, and the COAL baseline could stimulate and facilitate future research towards this important problem.}

 
\end{abstract}

\input{1_introduction.tex}
\input{2_related.tex}

\input{3.1_method.tex}
\input{4.1_experiments.tex}
\input{5_conclusion.tex}

\clearpage
\bibliography{egbib}
\bibliographystyle{splncs} 

\newpage
\appendix
\input{6_appendix.tex}

\end{document}

%% file: math_commands.tex
\usepackage{amsmath,amsfonts,bm}

\def\eqref#1{equation~\ref{#1}}

\def\1{\bm{1}}

\newcommand{\test}{\mathcal{D_{\mathrm{test}}}}

\DeclareMathAlphabet{\mathsfit}{\encodingdefault}{\sfdefault}{m}{sl}
\SetMathAlphabet{\mathsfit}{bold}{\encodingdefault}{\sfdefault}{bx}{n}

\def\gH{{\mathcal{H}}}

\newcommand{\R}{\mathbb{R}}

\DeclareMathOperator*{\argmin}{arg\,min}

%% file: 1_introduction.tex
\section{Introduction}
The success of deep learning models is highly dependent on the assumption that the training and testing data are \textit{i.i.d} and sampled from the same distribution. 
In reality, they are typically collected from different but related domains, leading to a phenomenon known as \textit{domain shift}~\cite{datashift_book2009}. To bridge the domain gap, Unsupervised Domain Adaptation (UDA) transfers the knowledge learned from a labeled source domain to an unlabeled target domain by statistical distribution alignment~\cite{long2015,ddc} or adversarial alignment~\cite{adda,DANN,MCD_2018}. Though recent UDA works have made great progress, \ssh{most of them are under the assumption that the prior label distributions of the two domains are identical.}
Denote the input data as $x$ and output labels as $y$, and let the source and target domain be characterized by probability distributions $p$ and $q$, respectively. The majority of \ssh{UDA} methods assume that the conditional label distribution is invariant ($p(y|x) = q(y|x)$), and only the \textit{\ssh{\SHIFTNAMEFULL}} ($p(x) \not= q(x)$) needs to be tackled, neglecting potential \textit{\ssh{\YSHIFTNAMEFULL}} ($p(y) \not= q(y)$) 
\footnote{\ssh{Different from some works \cite{pmlr-v80-lipton18a,azizzadenesheli2018regularized}, we do not assume $p(x|y)=q(x|y)$ for label shift.}}.  
\ssh{However, we claim that this assumption makes current UDA methods not applicable in the real world, for the following reasons:}
\ssh{\textbf{1)} this assumption hardly holds true in real applications, as \YSHIFTNAMEFULL~across domains is commonly seen in the real world}. For example, an autonomous driving system should be able to handle constantly changing frequencies of pedestrians and cars when adapting from a rural to a downtown area; or from a rainy to a sunny day.
\ssh{In addition, it is hard to guarantee $p(y)=q(y)$ without any information about $q(y)$ in the real world.}
\ssh{\textbf{2)} recent theoretical work~\cite{pmlr-v97-zhao19a} has demonstrated that if \YSHIFTNAME~exists, current UDA methods could lead to significant \textit{performance drop}. This is also empirically proved by our experiments.
\textbf{3)} we cannot check whether label shift exists in real applications. This prevents us from safely applying current UDA methods because we cannot predict the potential risk of performance drop}. \ssh{Therefore, we claim that an applicable UDA method must be able to handle \SHIFTNAMEFULL~and \YSHIFTNAMEFULL~at the same time.}


\begin{figure*}[t!]
    \includegraphics[width=1.0\linewidth]{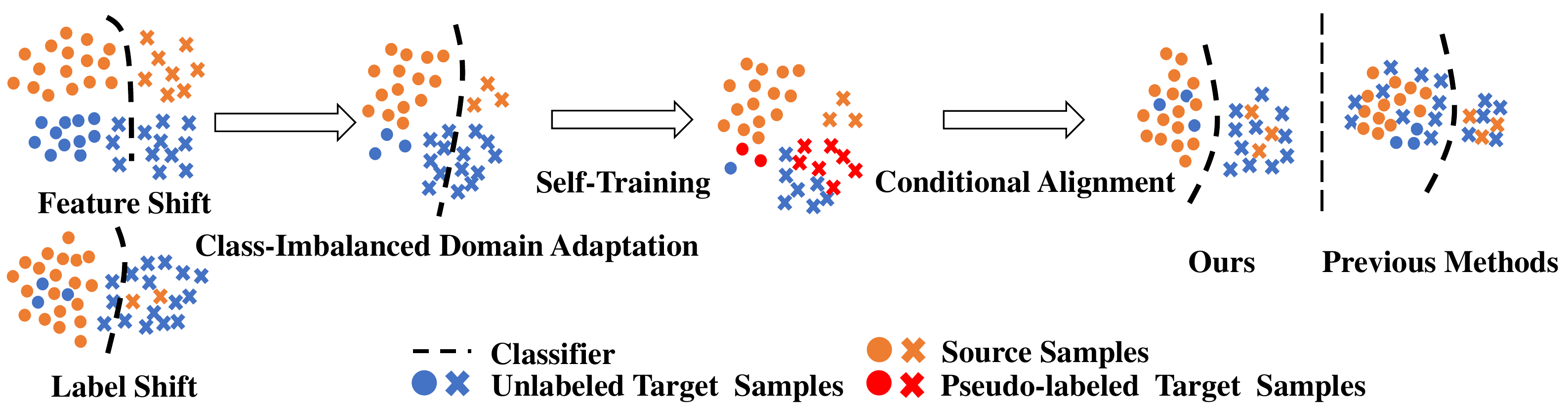}
    \caption{ 
    \ssh{We propose the \DANAMEFULL~setting, where we consider \SHIFTNAME~and \YSHIFTNAME~simultaneously. We provide the first empirical evaluation of this setting, showing that existing UDA methods are very fragile in the face of label shift. This is because learning marginal domain-invariant features will incorrectly align samples from different categories, leading to negative transfer. We propose an alternate, more robust approach that combines self-training and conditional feature alignment to tackle feature and label shift.}}
    \vspace{-.2in}
    \label{fig_GDA_overview}
\end{figure*}

To \ssh{formulate} the above problem, we propose
\ssh{\textbf{\textit{\DANAMEFULL}} (\textbf{\DANAME})}, a more challenging but practical domain adaptation setting where the {conditional \ssh{\SHIFTNAMEFULL}} and {\ssh{\YSHIFTNAMEFULL}} are required to be tackled \textit{simultaneously}. Specifically, \ssh{in addition to Covariate Shift assumption ($p(x) \not= q(x)$, $p(y|x) = q(y|x)$), we further assume $p(x|y) \not= q(x|y)$ and $p(y) \not= q(y)$.}
The main challenges of \DANAME~are: \textbf{1)}  {\ssh{\YSHIFTNAME}} hampers the effectiveness of mainstream domain adaptation methods that only marginally aligns feature distributions, \textbf{2)} aligning the conditional feature distributions ($p(x|y), q(x|y)$) is difficult in the presence of {\ssh{\YSHIFTNAME}}, and \textbf{3)} when data in one or both of the domains are unequally distributed across different categories, it is difficult to train an unbiased classifier. An overview of \DANAME~is shown in Figure \ref{fig_GDA_overview}.

\ssh{Aligned with our idea,} several works \cite{zhang2013domain,gong2016domain,wu2019domain} provide theoretical analyses \ssh{on domain adaptation with both feature and label shift. However, they do not provide sufficient empirical analysis of current UDA methods under this setting.} In addition, no practical algorithm that can solve real-world cross-domain problems has been proposed by these works.
\ssh{Therefore, although this problem has been known for years, most recent UDA methods are still not able to handle it. In this paper, we aim raise concerns and interests towards this important problem by taking one practical step. Firstly, we create \DANAME~benchmarks with 22 cross-domain tasks across 6 real-world image classification datasets. We believe this would facilitate future domain adaptation research towards robustly applicable methods. Secondly, we extensively evaluate 10 state-of-the-art domain adaptation methods to analysis how well \DANAME~is solved currently. We find most of these methods cannot handle \DANAME~well and often lead to negative transfer. Thirdly, towards a better solution, we provide a theoretically-motivated novel combination of existing ideas, which works well as a baseline for future research.}

\ssh{In this work, we visited domain adaptation methods in three categories}. Mainstream unsupervised domain adaptation aligns the feature distributions of two domains by methods that include minimizing the Maximum Mean Discrepancy~\cite{long2015,ddc}, aligning high-order moments~\cite{cmd,DomainNet}, or adversarial training~\cite{adda,DANN}. However, these models are limited when applied to the \DANAME~task as they only align the feature distribution, ignoring the issue of {\ssh{\YSHIFTNAME}} \cite{pmlr-v97-zhao19a}.
Another line of  works~\cite{pmlr-v80-lipton18a,azizzadenesheli2018regularized} assume that only {\ssh{\YSHIFTNAME}} exists ($p(y) \not= q(y)$) between two domains and the conditional feature distribution is invariant ($p(x|y) = q(x|y)$). These methods have achieved good performance when the data in both domains are sampled from the same feature distribution but under different label distributions. However, these models cannot handle the \DANAME~task as the feature distribution is not well aligned. \ssh{Recently, several works consider the domain adaptation problem where the categories of the source and target domain are not fully overlapped \cite{cao2018partial,ETN_2019_CVPR,UDA_2019_CVPR}. This setting can be seen as a special case of \DANAME~where for some class $i$ we have either $p(y=i)=0$ or $q(y=i)=0$. In our experiments, we showed that 8 out of 10 methods we evaluated on \DANAME~tasks frequently lead to negative transfer (produce worse performance than no-adaptation baseline), while the rest methods only leads to limited improvement over the baseline on average. This limited performance showed that current UDA methods are not robust enough to be practically applied, and  motivated us to reconsider the solution to the \DANAME~problem.}

We postulate that it is essential to align the conditional feature distributions as well as the label distributions to tackle the \DANAME~task. In this work, we address \DANAME~with \ssh{feature distribution} and \ssh{label distribution} \textbf{CO-ALignment} (\textbf{\MODELNAME}). \ssh{Specifically, to deal with feature shift and label shift in an unified way, we proposed a simple baseline method that combines the ideas of \textit{prototype-based conditional distribution alignment} \cite{saito2019semi} and \textit{class-balanced self-training} \cite{zou2018unsupervised}}.
First, to tackle \ssh{\SHIFTNAME} in the context of \ssh{\YSHIFTNAME}, it is essential to align the conditional rather than marginal feature distributions, to avoid the negative transfer effects caused by matching marginal feature distributions~\cite{pmlr-v97-zhao19a} (illustrated in Figure~\ref{fig_GDA_overview}). To this end, we \ssh{use} a prototype-based method to align the conditional feature distributions of the two domains. The \textit{source} prototypes are computed by learning a similarity-based classifier, \ssh{which are moved towards the \textit{target} domain} with a minimax entropy algorithm~\cite{saito2019semi}.
Second, we align the label distributions in the context of \ssh{\SHIFTNAME}~by \ssh{training the classifier with estimated} target label distribution through a class-balanced self-training method \cite{zou2018unsupervised}.
We incorporate the above \ssh{feature distribution} and \ssh{label distribution} alignment into an end-to-end deep learning framework, as illustrated in Figure~\ref{fig_method}.
Comprehensive experiments on standard cross-domain recognition benchmarks demonstrate that \MODELNAME~achieves significant improvements over the state-of-the-art methods on the task of \DANAME. 

The main contributions of this paper are highlighted as follows: \textbf{1)} to the best of our knowledge, we provide the first set of benchmarks and practical solution for domain adaptation under joint \ssh{feature and label shift} in deep learning, \ssh{which is important for real-world applications}; \textbf{2)} we deliver extensive experiments to demonstrate that state-of-the-art methods \textit{fail} to align \ssh{feature distribution}~in the presence of \ssh{label distribution}, or \text{vise versa}; \textbf{3)} \ssh{we propose a simple yet effective feature and label distribution CO-ALignment (\MODELNAME) framework, which could be a useful baseline for future research towards practical domain adaptation. We believe the provided benchmarks, empirical analysis and the baseline model could trigger future research works towards more practical domain adaptation.}

%% file: 2_related.tex
\section{Related Work}

\textbf{Domain Adaptation for Feature Shift} Domain adaptation aims to transfer the knowledge learned from one or more source domains to a target domain. Recently, many unsupervised domain adaptation methods have been proposed. These methods can be taxonomically divided into three categories~\cite{DA_Survey}. The first category is the discrepancy-based approach, which leverages different 
measures to align the marginal feature distributions between source and target domains. Commonly used measures include Maximum Mean Discrepancy (MMD)~\cite{JAN,tzeng2014deep}, $\gH$-divergence~\cite{ben2010theory}, Kullback-Leibler (KL) divergence~\cite{zhuang2015supervised}, and Wasserstein distance~\cite{lee2017minimax,shen2017wasserstein}. The second category is the adversarial-based approach~\cite{adda,cogan,Peng2019DomainAL} which uses a domain discriminator to encourage domain confusion via an adversarial objective. The third category is the reconstruction-based approach. Data are reconstructed in the new domain by an encoder-decoder~\cite{bousmalis2016domain,ghifary2016deep} or a GAN discriminator, such as dual-GAN~\cite{yi2017dualgan}, cycle-GAN~\cite{CycleGAN2017}, disco-GAN~\cite{kim2017learning}, and CyCADA~\cite{hoffman2017cycada}. However, these methods mainly consider aligning the marginal distributions to decrease~\ssh{\SHIFTNAME}, neglecting~\ssh{\YSHIFTNAME}. To the best of our knowledge, we are the first the propose an end-to-end deep model to tackle both of the two domain shifts between the source and target domains. 

\noindent \textbf{Domain Adaptation for \ssh{Label Shift}} Despite its wide applicability,  learning under \ssh{\YSHIFTNAME} remains under-explored. Existing works tackle this challenge by importance reweighting or target distribution estimation. Specifically, \cite{zhang2013domain} exploit importance reweighting to enhance  knowledge transfer under \ssh{\YSHIFTNAME}. Recently, \cite{lipton2018detecting} introduce a test distribution estimator to detect and correct for \ssh{\YSHIFTNAME}. These methods assume that the source and target domains share the same feature distributions and only differ in the marginal label distribution. In this work, we explore transfer learning between domains under \ssh{\YSHIFTNAME~and \YSHIFTNAME~simultaneously}.
As a special case of \ssh{\YSHIFTNAME}, some works consider the domain adaptation problem where the categories in the source domain and target domain are not fully overlapped. \cite{panareda2017open} propose \textit{open set domain adaptation} where the class set in the source domain is a proper subset of that of the target domain. Conversely, \cite{cao2018partial} introduce \textit{partial domain adaptation} where the class set of the source domain is a proper superset of that of the target domain. 
In this direction, ~\cite{zhao2019learning} introduce a theoretical analysis to show that only learning domain-invariant features is not sufficient to solve domain adaptation task when the label priors are not aligned. In a related work, \cite{wu2019domain} propose asymmetrically-relaxed distribution alignment to overcome the limitations of standard domain adaptation algorithms which aims to extract domain-invariant representations.

\noindent \textbf{Domain adaptation with self-training}
In domain adaptation, self-training methods are often utilized to compensate for the lack of categorical information in the target domain. The intuition is to assign pseudo-labels to unlabeled samples based on the predictions of one or more classifiers.  
\cite{saito2017asymmetric} leverage an asymmetric tri-training strategy to assign pseudo-labels to the unlabeled target domain.~\cite{xie2018learning} propose to assign pseudo-labels to all target samples and use them to achieve semantic alignment across domains.

Recently, ~\cite{chen2019progressive} propose to progressively label the target samples and align the prototypes of source domain and target domain to achieve  domain alignment. 
However, to the best of our knowledge, self-training has not been applied for DA with \YSHIFTNAME.

%% file: 3.1_method.tex
\section {CO-ALignment of \ssh{Feature and Label Distribution}}

In \DANAMEFULL, we are given a \textit{source} domain $\mathcal{D_S}=\{(x_i^s, y_i^s)_{i=1}^{N_s}\}$ with $N_s$ labeled examples, and a \textit{target} domain $\mathcal{D_T}=\{(x_i^t)_{i=1}^{N_t}\}$ with $N_t$ unlabeled examples. We assume that $p(y|x) = q(y|x)$ but $p(x|y) \not= q(x|y)$, $p(x) \not= q(x)$, and $p(y) \not= q(y)$. We aim to construct an end-to-end deep neural network which is able to transfer the knowledge learned from $\mathcal{D_S}$ to $\mathcal{D_T}$, and train a classifier $y = \theta(x)$ which can minimize task risk in target domain $\epsilon_T(\theta) = \text{Pr}_{(x,y)\sim q}[\theta(x)\not=y]$. 

\begin{figure*}[t!]
    \includegraphics[width=1.02\linewidth]{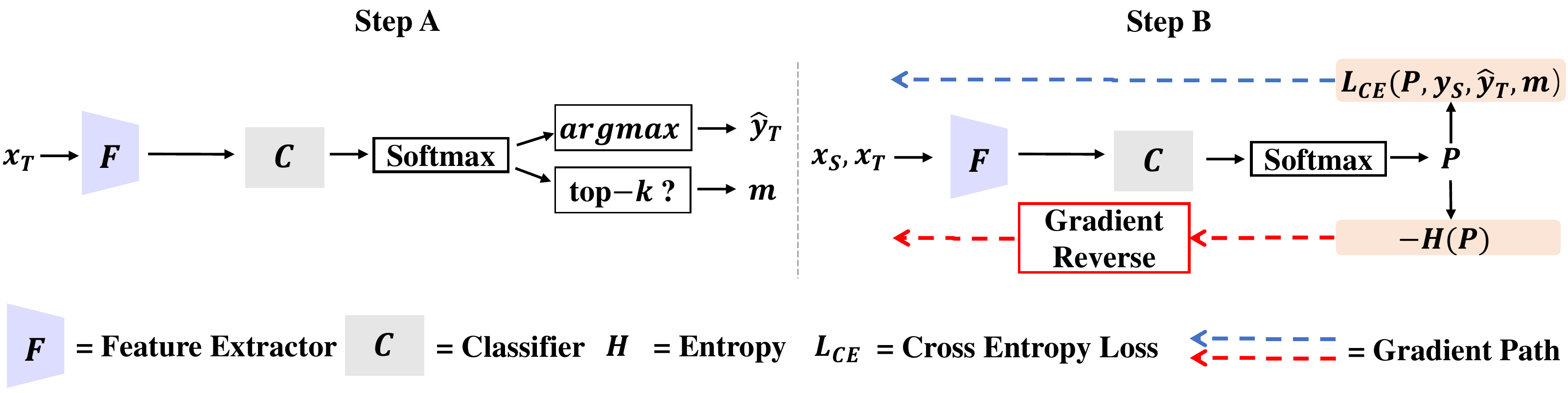}
    \caption{\textbf{Overview of the proposed COAL model}. Our model is trained iteratively between two steps. In step A, we forward the target samples through our model to generate the pseudo labels and mask. In step B, we train our models by \textit{self-training} with the pseudo-labeled target samples to align the label \ssh{distributions}, and \textit{prototype-based conditional alignment} with the minimax entropy.}
    \label{fig_method}
    \vspace{-0.2in}
\end{figure*}

Previous works either focus on aligning the marginal \ssh{feature distributions} ~\cite{long2015,adda} or aligning the label \ssh{distributions}~\cite{lipton2018detecting}. These approaches are not able to fully tackle \DANAME~as they only align one \ssh{of the two marginal distributions}. \ssh{Motivated by theoretical analysis, in this work we propose to tackle \DANAME~with feature distribution and label distribution CO-ALignment. To this end, we combine the ideas of \textit{prototype-based conditional alignment} \cite{saito2019semi} and \textit{class-balanced self-training} \cite{zou2018unsupervised} to tackle feature and label shift respectively. An overview of COAL is shown in Figure \ref{fig_method}. }

\subsection{Theoretical Motivations}
\noindent \textbf{Conditional Feature Alignment}
According to \cite{zhao2019learning}, the target error in domain adaptation is bounded by three terms: 1) source error, 2) the discrepancy between the marginal distributions and 3) the distance between the source and target optimal labeling functions. Denote $h \in \mathcal{H}$ as the hypothesis, $\epsilon_S(\cdot)$ and $\epsilon_T(\cdot)$ as the expected error of a labeling function on source and target domain, and $f_S$ and $f_T$ as the optimal labeling functions in the source and target domain. Then, we have:
\begin{equation}
    \label{theory1}
    \epsilon_T(h) \leq \epsilon_S(h) + d_{\hat{\mathcal{H}}} (\mathcal{D}_S, \mathcal{D}_T) + \min\{\epsilon_S(f_T), \epsilon_T(f_S) \},
\end{equation}
where $d_{\hat{\mathcal{H}}}$ denote the discrepancy of the marginal distributions \cite{zhao2019learning}. The bound demonstrates that the optimal labeling functions $f_S$ and $f_T$ need to generalize well in both domains, such that the term $\min\{\epsilon_S(f_T), \epsilon_T(f_S) \}$ can be bounded. Conventional domain adaptation approaches which only align marginal feature distribution cannot guarantee that $\min\{\epsilon_S(f_T), \epsilon_T(f_S)\}$ is minimized. This motivates us to align the conditional feature distribution, \textit{i.e.} $p(x|y)$ and $q(x|y)$. 

\noindent \textbf{Class-balanced Self-training}
Theorem 4.3 in~\cite{zhao2019learning} indicates that the target error $\epsilon_T(h)$ can not be minimized if we only align the \ssh{feature distributions} and neglect \ssh{the shift in label distribution}. Denote $d_{JS}$ as the Jensen-Shannon(JS) \ssh{distance} between two distributions, \cite{zhao2019learning} propose:
\begin{equation}
    \label{theory2}
    \epsilon_S(h) + \epsilon_T(h) \geq \frac{1}{2} (d_{JS} (p(y), q(y)) - d_{JS}(p(x), q(x))) ^ 2
\end{equation}
This theorem demonstrates that when the divergence between label distributions $d_{JS} (p(y), q(y))$ is significant, minimizing the divergence between marginal distributions $d_{JS}(p(x), q(x))$ and the source task error $\epsilon_S(h)$ will enlarge the target task error $\epsilon_T(h)$. Motivated by this, we propose to \ssh{estimate and} align the empirical label \ssh{distributions} with a self-training algorithm. 

\subsection{Prototype-based Conditional Alignment for \ssh{Feature Shift}}
The mainstream idea in feature-shift oriented methods is to learn domain-invariant features by aligning the marginal feature distributions, which was proved to be inferior in the presence of label shift~\cite{pmlr-v97-zhao19a}. 
Instead, \ssh{inspired by \cite{saito2019semi}}, we align the conditional feature distributions $p(x|y)$ and $q(x|y)$. To this end, we leverage a \textit{similarity-based classifier} to estimate $p(x|y)$, and a minimax entropy algorithm to 
align it with $q(x|y)$. We achieve conditional \ssh{feature distribution} alignment by aligning the source and target prototypes in an adversarial process.


\noindent \textbf{Similarity-based Classifier} The architecture of our \MODELNAME~model contains a feature extractor $F$ and a similarity-based classifier $C$. Prototype-based classifiers perform well in few-shot learning settings~\cite{chen2019closerfewshot}, which motivates us to adopt them since in label-shift settings some categories can have low frequencies.
Specifically, $C$ is composed of a weight matrix $\textbf{W}\in \R^{d\times c}$ and a temperature parameter $T$, where $d$ is the dimension of feature generated by $F$, and $c$ is the total number of classes. Denote $\textbf{W}$ as $[\textbf{w}_1,\textbf{w}_2,..., \textbf{w}_c]$, this matrix can be seen
as $c$ $d$-dimension vectors, one for each category. For each input feature $F(x)$, we compute its similarity with the $i_{th}$ weight vector as $s_i = \frac{F(x)\textbf{w}_i}{T \left \| F(x)\right \|}$. Then, we compute the probability of the sample being labeled as class $i$ by $h_i(x)=\sigma(\frac{F(x)\textbf{w}_i}{T \left \| F(x)\right \|})$,   normalizing over all the classes. Finally, we can compute the prototype-based classification loss for $\mathcal{D_S}$ with standard cross-entropy loss: 
\begin{equation}
\label{loss:supervise}
\mathcal{L}_{SC} = \mathbb{E}_{(x,y)\in\mathcal{D}_S} \mathcal{L}_{ce}(h(x), y)    
\end{equation}

The intuition behind this loss is that the higher the confidence of sample $x$ being classified as class $i$, the closer the embedding of $x$ is to $\textbf{w}_i$. Hence, when optimizing Equation \ref{loss:supervise}, we are reducing the intra-class variation by pushing the embedding of each sample $x$ closer to its corresponding weight vector in $\textbf{W}$. In this way, $\textbf{w}_i$ can be seen as a representative data point (prototype) for $p(x|y=i)$.

\noindent \textbf{Conditional Alignment by Minimax Entropy} Due to the lack of categorical information in the target domain, it is infeasible to utilize Equation \ref{loss:supervise} to obtain target prototypes. \ssh{Following \cite{saito2019semi}}, we tackle this problem by 1) moving each source prototype 
to be closer to its nearby target samples, and 2) clustering target samples around this moved prototype. 
We achieve these two objectives jointly by entropy minimax learning. Specifically, for each sample $x^t \in \mathcal{D_T}$ fed into the network, we can compute the mean entropy of the classifier's output by 
\begin{equation}
\label{equ:coal_entropy}
\mathcal{L}_{H}= \mathbb{E}_{x\in\mathcal{D_T}} H(x)= -\mathbb{E}_{x\in\mathcal{D_T}}\sum_{i=1}^{c}h_i(x)\log{h_i(x)}.
\end{equation}
Larger $H(x)$ indicates that \ssh{sample $x$ is} similar to all the weight vectors (prototypes) of $C$. We achieve conditional feature distributions alignment by aligning the source and target prototypes in an adversarial process: (1) we train $C$ to \textit{maximize} $\mathcal{L}_{H}$, aiming to move the prototypes from the source samples towards the neighboring target samples; (2) we train $F$ to \textit{minimize} $\mathcal{L}_{H}$, aiming to make the embedding of target samples closer to their nearby prototypes. By training with these two objectives as a min-max game between $C$ and $F$, we can align source and target prototypes. Specifically, we add a gradient-reverse layer \cite{DANN} between $C$ and $F$ to flip the sign of gradient.

\subsection{\ssh{Class-balanced Self-training for Label Shift}}\label{method:label_shift}
As the source label distribution $p(y)$ is different from that of the target $q(y)$, it is not guaranteed that the classifier $C$ which has low risk on $\mathcal{D_S}$ will have low error on $\mathcal{D_T}$. Intuitively, if the classifier is trained with imbalanced source data, the decision boundary will be dominated by the most frequent categories in the training data, leading to a classifier \ssh{biased towards source label distribution}. When the classifier is applied to target domain with a different label distribution, its accuracy will degrade as it is highly biased \ssh{towards the source domain}. To tackle this problem, we \ssh{use the method in \cite{zou2018unsupervised} to} employ \textit{self-training} to estimate the target label distribution and refine the decision boundary. In addition, we leverage \textit{balanced sampling} of the source data to further \ssh{facilitate this process.}

\noindent \textbf{Self-training} 
\label{method_PL}
In order to refine the decision boundary, we propose to estimate the target label distribution with self-training. We assign pseudo labels $\hat{y}$ to all the target samples according to the output the classifier $C$. As we are also aligning the conditional \ssh{feature distributions} ~($p(x|y)$ and $q(x|y)$), we assume that the distribution \ssh{of high-confidence pseudo labels} $q(\hat{y})$ \ssh{can be used as an approximation} of the real label distribution $q(y)$ for the target domain. \ssh{Training $C$ with these pseudo-labeled target samples under approximated target label distribution, we are able to reduce the negative effect of label shift.}

\ssh{To obtain high-confidence pseudo labels, }for each category, we select  top $k\%$ of the target samples with the highest confidence scores belonging to that category. We utilize the highest probability in $h(x)$ as the classifier's confidence on sample $x$. Specifically, for each pseudo-labeled sample $(x,\hat{y})$, we set its selection mask $m=1$ if $h(x)$ is among the top $k\%$ of all the target samples with the same pseduo-label, otherwise $m=0$.
Denote the pseudo-labeled target set as $\hat{\mathcal{D}}_T=\{(x^t_i, \hat{y}^t_i, m_i)_{i=1}^{N_t}\}$, we leverage the input and pseudo labels from $\hat{\mathcal{D}}_T$ to train the classifier $C$, aiming to refine the decision boundary with target label distribution. The total loss function for classification is:
\begin{equation}
\label{loss:supervise_PL}
\mathcal{L}_{ST} =  \mathcal{L}_{SC} + \mathbb{E}_{(x,\hat{y},m)\in \hat{\mathcal{D}}_T}  \mathcal{L}_{ce}(h(x), \hat{y}) \cdot m
\end{equation}
where $\hat{y}$ indicates the pseudo labels and $m$ indicates selection masks. In our approach, we choose the top $k\%$ of the highest confidence target samples \textit{within} each category, instead of universally. This is crucial to estimate the real target label distribution, otherwise, the easy-to-transfer categories will dominate $\hat{\mathcal{D}}_T$, leading to inaccurate estimation of the target label distribution \cite{zou2018unsupervised}. \ssh{As training processes, we are able to obtain pseudo labels with higher accuracy. Therefore, we increase $k$ by $k_{step}$ after each epoch until it reaches a threshold $k_{max}$. Typically, we initialize $k$ with $k_0=5$, and set $k_{step}=5$, $k_{max}=30$.}

\noindent \textbf{Balanced Sampling of Source Data} 
When coping with label shift, the label distribution of the source domain could be highly imbalanced. A classifier trained on imbalanced categories will make highly-biased predictions for the samples from the target domain \cite{He:2009:LID:1591901.1592322}. This effect also hinders the self-training process discussed above, as the label distribution estimation will  also be biased. To tackle these problems, we apply a balanced mini-batch sampler to generate training data from the source domain and ensure that each source mini-batch contains roughly the same number of samples for each category. 

\subsection{Training Process}
In this section, we combine the above ideas into an end-to-end training pipeline. Denote $\alpha$ as the trade-off between classifier training and feature distribution alignment, we first define the adaptive learning objective as follows:

\begin{equation}
\label{loss:minmax}
\begin{split}
\hat{C} = \argmin_{C} \mathcal{L}_{ST} - \alpha \mathcal{L}_H , \ \ \ \ \ \ \ 
\hat{F} = \argmin_{F} \mathcal{L}_{ST} + \alpha \mathcal{L}_H .
\end{split}
\end{equation}

Given input samples from source domain $\mathcal{D}_S$ and target domain $\mathcal{D}_T$, we first pretrain our network $F$ and $C$ with only labeled data $\mathcal{D}_S$. Then, we iterate between \textbf{pseudo-label assignment} (step A) and \textbf{adaptive learning} (step B). 
\ssh{We update the pseudo labels in each epoch as we obtain better feature representations from adaptive learning, which leads to more accurate pseudo labels. On the other hand, better pseudo labels could also facilitate adaptive learning in the next epoch.}
This process continues until convergence or reaching the maximum number of iterations. An overview of it is shown in Figure \ref{fig_method}.

%% file: 4.1_experiments.tex
\section{Experiments}
\label{exp:total}
\ssh{In this section, we first construct the \DANAME~benchmarks with 26 cross-domain adaptation tasks based on 4 \textbf{Digits} datasets, \textbf{Office-Home}~\cite{OfficeHome} and \textbf{DomainNet}~\cite{DomainNet}. Then we evaluate and analysis 10 representative state-of-the-art domain adaptation methods as well as our \MODELNAME~baseline. Finally, we provide additional analysis experiments to further explore the \DANAME~problem.}

\subsection{\ssh{\DANAMEFULL~Benchmark}}

\begin{figure*}[t]
    \centering
    \begin{subfigure}{\linewidth}    \includegraphics[width=\linewidth]{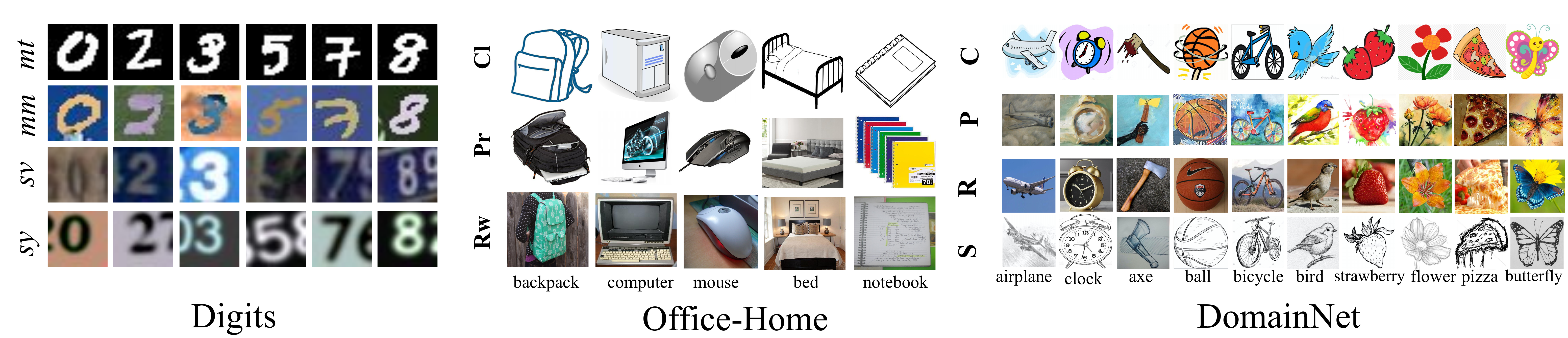}
    \caption{Sample images of the datasets we use in our experiments.}
    \label{fig:dataset_overview}
    \end{subfigure}

    \centering
    \begin{subfigure}[b]{0.5\textwidth}
        \centering
        \includegraphics[scale=0.35]{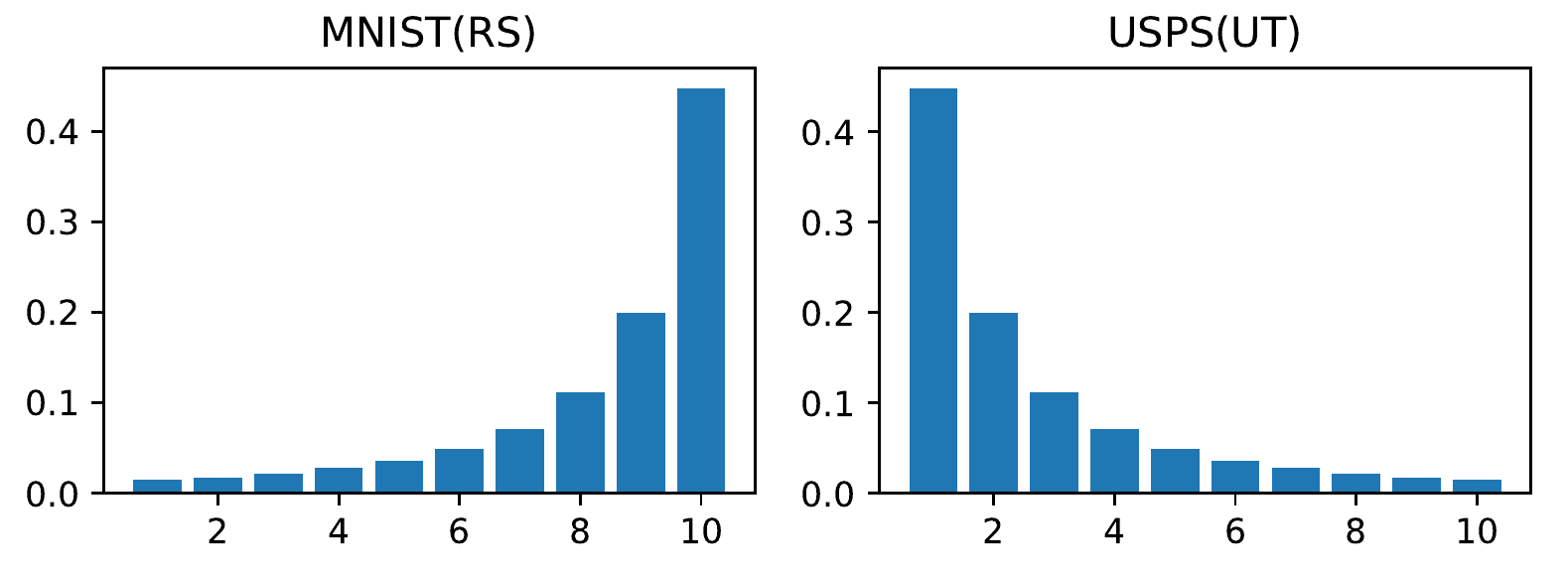}
        \caption{\textbf{Digits} \ssh{\YSHIFTNAME}}
    \end{subfigure}%
    ~ 
    \begin{subfigure}[b]{0.5\textwidth}
        \centering
        \includegraphics[scale=0.35]{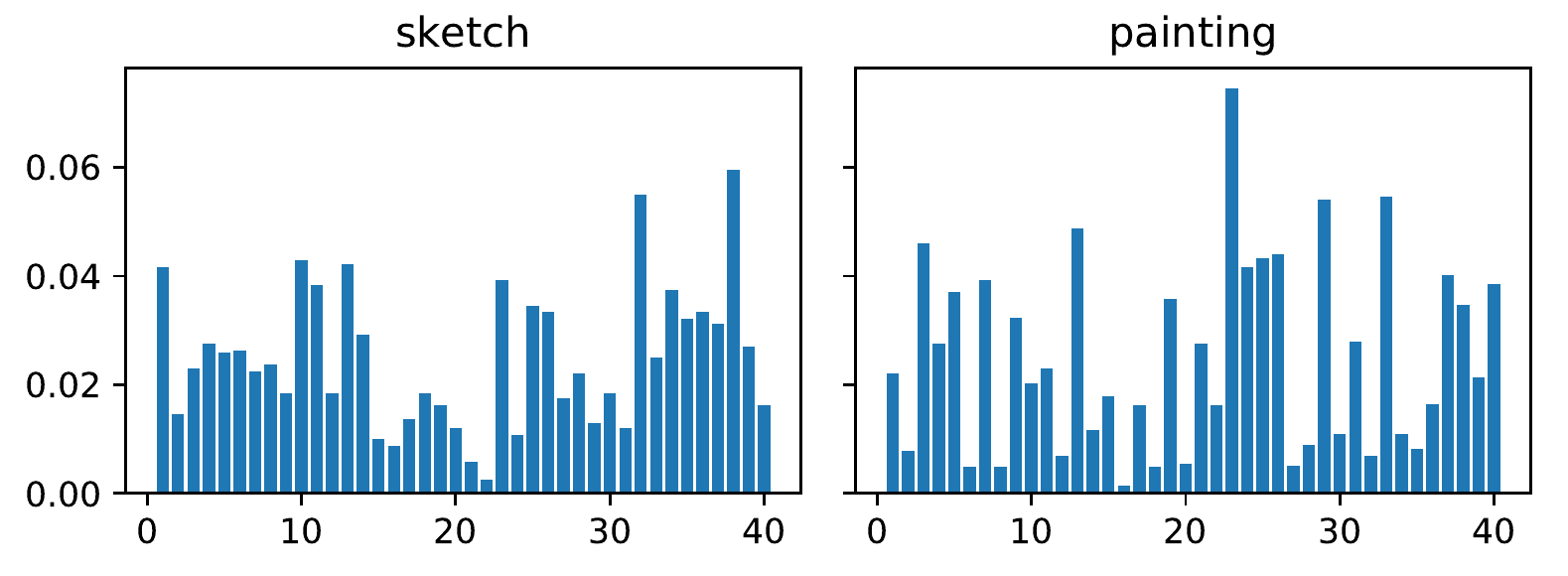}
        \caption{\textbf{DomainNet} \ssh{\YSHIFTNAME}}
    \end{subfigure}
    \caption{(\textbf{a}): Image examples from Digits, Office-Home~\cite{OfficeHome}, and DomainNet~\cite{DomainNet}. (\textbf{b}): illustrations of \textbf{R}eversely-unbalanced \textbf{S}ource (\textbf{RS}) and \textbf{U}nbalanced \textbf{T}arget (\textbf{UT}) distribution in MNIST$\rightarrow$USPS task. (\textbf{c}): Natural \ssh{\YSHIFTNAME} of DomainNet.}
    \label{fig:distribution}
\end{figure*}

\noindent \textbf{\ssh{Domain Shift Protocol.}}
\ssh{Because the images use are already collected from separate feature domains, we only create label shift for each cross-domain task.} To create \ssh{\YSHIFTNAME} between source and target domains, we sub-sample the current datasets with \textbf{R}eversely-unbalanced \textbf{S}ource and \textbf{U}nbalanced \textbf{T}arget (\textbf{RS-UT}) protocol. In this setting, both the source and target domains have unbalanced label distribution, while the label distribution of the source domain is a reversed version of that of the target domain. \ssh{Following \cite{openlongtailrecognition}, the unbalanced label distribution is created by sampling from a Paredo distribution \cite{Pareto}}. An illustration of this setting can be found in Figure \ref{fig:distribution}(b). We refer our reader to supplementary material for detailed data splits and creation process.

\noindent \textbf{Digits.} We select four digits datasets: MNIST \cite{MNIST}, USPS \cite{usps}, SVHN \cite{svhn} and Synthetic Digits (SYN) \cite{ganin2015unsupervised} and regard each of them as a separate domain.
In this work, we investigate four domain adaptation tasks: \textbf{MNIST} $\rightarrow$ \textbf{USPS}, \textbf{USPS} $\rightarrow$ \textbf{MNIST}, \textbf{SVHN} $\rightarrow$ \textbf{MNIST}, and \textbf{SYN} $\rightarrow$ \textbf{MNIST}.
    
\noindent \textbf{Office-Home} \cite{OfficeHome} is a dataset collected in office and home environment with 65 object classes and four domains: Real World (\textbf{Rw}), Clipart (\textbf{Cl}), Product (\textbf{Pr}), Art (\textbf{Ar}). Since the ``Art'' domain is too small to sample an imbalanced subset, we focus on the remaining domains and explore all the six adaptation tasks.

\noindent \textbf{DomainNet} \cite{DomainNet} is a large-scale testbed for
domain adaptation, which contains six domains with about 0.6 million images distributed among 345 classes. Since some domains and classes contains many mislabeled outliers, we select 40 commonly-seen classes from four domains: Real (\textbf{R}), Clipart (\textbf{C}), Painting (\textbf{P}), Sketch (\textbf{S}). Different from the two datasets above, the existed \ssh{\YSHIFTNAME} in DomainNet is significant enough, as illustrated in Figure \ref{fig:distribution}(c).
\ssh{Therefore, we use the original label distributions without sub-sampling for this dataset.}

\noindent \textbf{Evaluated Methods.} \ssh{To form a comprehensive empirical analysis, we evaluated recent domain adaptation methods from three categories, including \textbf{1)} conventional UDA methods that only aligns \ssh{feature distribution}: \textbf{DAN}~\cite{long2015}, \textbf{JAN}~\cite{JAN}, \textbf{DANN}~\cite{DANN},  \textbf{MCD}~\cite{MCD_2018} and \textbf{BSP}~\cite{BSP}; \textbf{2)} method that only aligns \ssh{label distribution}: \textbf{BBSE}\cite{lipton2018detecting}; \textbf{3)} methods that align feature distribution while assuming non-overlapping label spaces: \textbf{PADA}~\cite{cao2018partial}, \textbf{ETN}~\cite{ETN_2019_CVPR} and \textbf{UAN}~\cite{UDA_2019_CVPR}. We also evaluated \textbf{FDANN} \cite{wu2019domain}, which relaxes the feature distribution alignment objective in DANN to deal with potential label shift.}

\input{4.2_DigitResult.tex}
\input{4.3_OfficeHomeResult.tex}
\input{4.5_DomainNetResult.tex}
\noindent \textbf{Implementation Details.} We implement all our experiments in Pytorch platform. \ssh{We used the official implements for all the evaluated methods except for DANN \cite{DANN}, BBSE \cite{lipton2018detecting} and FDANN \cite{wu2019domain}, which are reproduced by ourselves. For fair comparison, we use the same backbone networks for all the methods. Specifically, for the Digits dataset, we adopt the network architecture proposed by \cite{MCD_2018}. For the other two datasets, we utilize ResNet-50 \cite{He2015DeepRL} as our backbone network, and replace the last fully-connected layer with a randomly initialized N-way classifier layer (for N categories). For all the compared methods, we select their hyper-parameters on the validation set of P$\rightarrow$C task of DomainNet. We refer our reader to supplementary material for code and parameters of COAL.}

\noindent \textbf{Evaluation metric.} When the target domain is highly unbalanced, conventional overall average accuracy that treats every class uniformly is not an appropriate performance metric \cite{he2008learning}. Therefore, we follow \cite{8353718} to use the \textit{per-class} mean accuracy in our main results. Formally, we denote $S_i = \frac{n_{(i,i)}}{n_i}$ as the accuracy for class $i$, where $n_{(i,j)}$ represents the number of class $i$ samples labeled as class $j$, and $n_i = \sum^c_{j=1} n_{(i,j)}$ represents the number of samples in class $i$. Then, the per-class mean accuracy is computed as $S=\frac{1}{c}\sum^c_{i=1}S_i$.

\subsection{Result Analysis}
We first show the experimental results on Digits datasets in Table~\ref{tab:digit}.
From the results, we can make the following observations:
\ssh{\textbf{(1)} Most current domain adaptation methods cannot solve \DANAME~well. On average, 8 of the 10 evaluated domain adaptation methods perform \textit{worse} than the source-only baselines, leading to negative transfer. This result confirmed the theoretical analysis that only aligning marginal feature distribution leads to performance drop under \DANAME~\cite{pmlr-v97-zhao19a}. 
\textbf{(2)} Method that achieve better results on conventional UDA benchmarks does not lead to better results on \DANAME~problem. For example, although MCD is shown to significantly outperform DAN and DANN on several conventional domain adaptation benchmarks \cite{MCD_2018}, its performance is inferior to these older methods in our experiment. We argue that this is because these newer methods achieve better marginal feature distribution alignment, which yet leads to worse performance under label shift. \textbf{(3)} Our COAL baseline achieves \textbf{84.33}\% average accuracy across four experimental setting, outperforming the best-performing method by \textbf{8.4}\%. This result demonstrate that aligning only the feature distributions or only the label distributions can not fully tackle \DANAME~task. In contrast, our framework co-aligns the conditional feature distributions and label distributions.}

Next, we show the experimental results on more challenging real-object \\ datasets, \textit{i.e.}, Office-Home and DomainNet, in Table \ref{tab:OH} and Table \ref{tab:DM}, respectively. In Office-Home experiments, we can also have the above observations.
For example, we observe that 7 out of 10 methods lead to negative transfer, which is consistent with the results on Digits dataset. 
Our COAL framework achieves \ssh{\textbf{58.87}}\% average accuracy across the six \DANAME~tasks, outperforming other evaluated methods, and has \ssh{\textbf{7.32}}\% improvement from the source-only result. 

In DomainNet experiments, \ssh{due to smaller degree of label shift, most evaluated methods could outperform the source-only baseline. However, we still observe the negative influence of label shift.} \ssh{First, we observe inferior performance of newer methods to older methods. For example, DANN outperformed MCD by 9.04\%, due to the negative effect of stronger marginal alignment in MCD. Moreover, our model get \ssh{\textbf{75.89}\%} average accuracy across the 12 tasks, outperforming all the compared baselines. This shows the effectiveness of feature and label distribution co-alignment in this dataset.} 
Furthermore, we carefully tuned the hyper-parameters for the evaluated domain adaptation methods to have \ssh{weaker feature distribution alignment} \footnote{Please refer to supplementary material for details.}. If we directly apply the parameters set by the authors, many of these models have much worse performance. 

\input{4.6_BalancedSamplingResult.tex}

\subsection{Analysis}
\label{exp:analysis} 
\noindent \textbf{Effect of Source Balanced Sampler.}
Source balanced samplers described in Section \ref{method:label_shift} can help us tackle the biased-classifier problem caused by the imbalanced data distribution of source domain. A significant performance boost can be observed  after applying the balanced sampler for our \MODELNAME~model, as well as the compared baselines. 
In this section, we specifically show the effect of using source balanced samplers. We show in Table \ref{tab:balanced} the performance of several methods with and without source balanced samplers on 5 adaptation tasks from multiple datasets. We observe that for 20 of the total 25 tasks (5 models on 5 adaptation tasks), using source balanced samplers will significantly improve the domain adaptation performance. 
These results show the effectiveness of having a source balanced sampler when tackling \DANAME~task.

\noindent \textbf{Ablation Study.}
Our \MODELNAME~method has mainly two objectives: 1) alignment of conditional feature distribution
\ssh{$\mathcal{L}_{ST}$}
and 2) alignment of label distribution 
\ssh{$\mathcal{L}_{H}$}
To show the importance of these two objectives in \DANAME, we show the performance of our method without each of these objectives respectively on multiple tasks. The results in Table \ref{tab:ablation} showed the importance of both objectives. For example, for USPS $\rightarrow$ MNIST, if we remove the conditional feature distribution alignment objective, the accuracy of our model will drop by $2.6\%$. Similarly, if we remove the label distribution alignment objective, the accuracy will drop by $2.9\%$. These results demonstrate that both the alignment of conditional feature distribution and label distribution are important to \DANAME~task.

\begin{figure}[t]
    \bigskip
    \begin{subfigure}{.18\linewidth}
        \centering
        \includegraphics[scale=0.16]{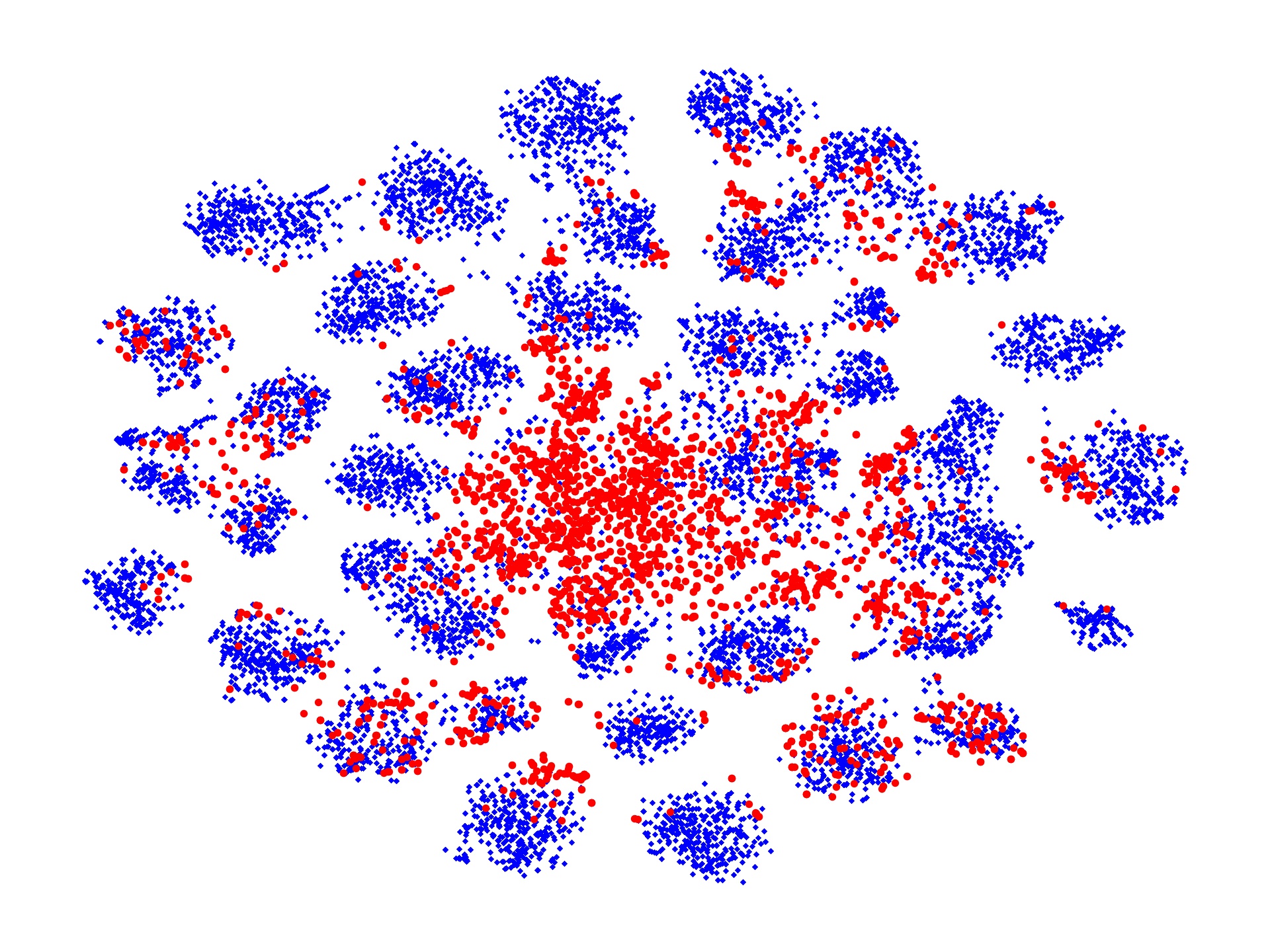}
        \caption{Source Only}
    \end{subfigure}
        \hfill
    \begin{subfigure}{.18\linewidth}
        \centering
        \includegraphics[scale=0.16]{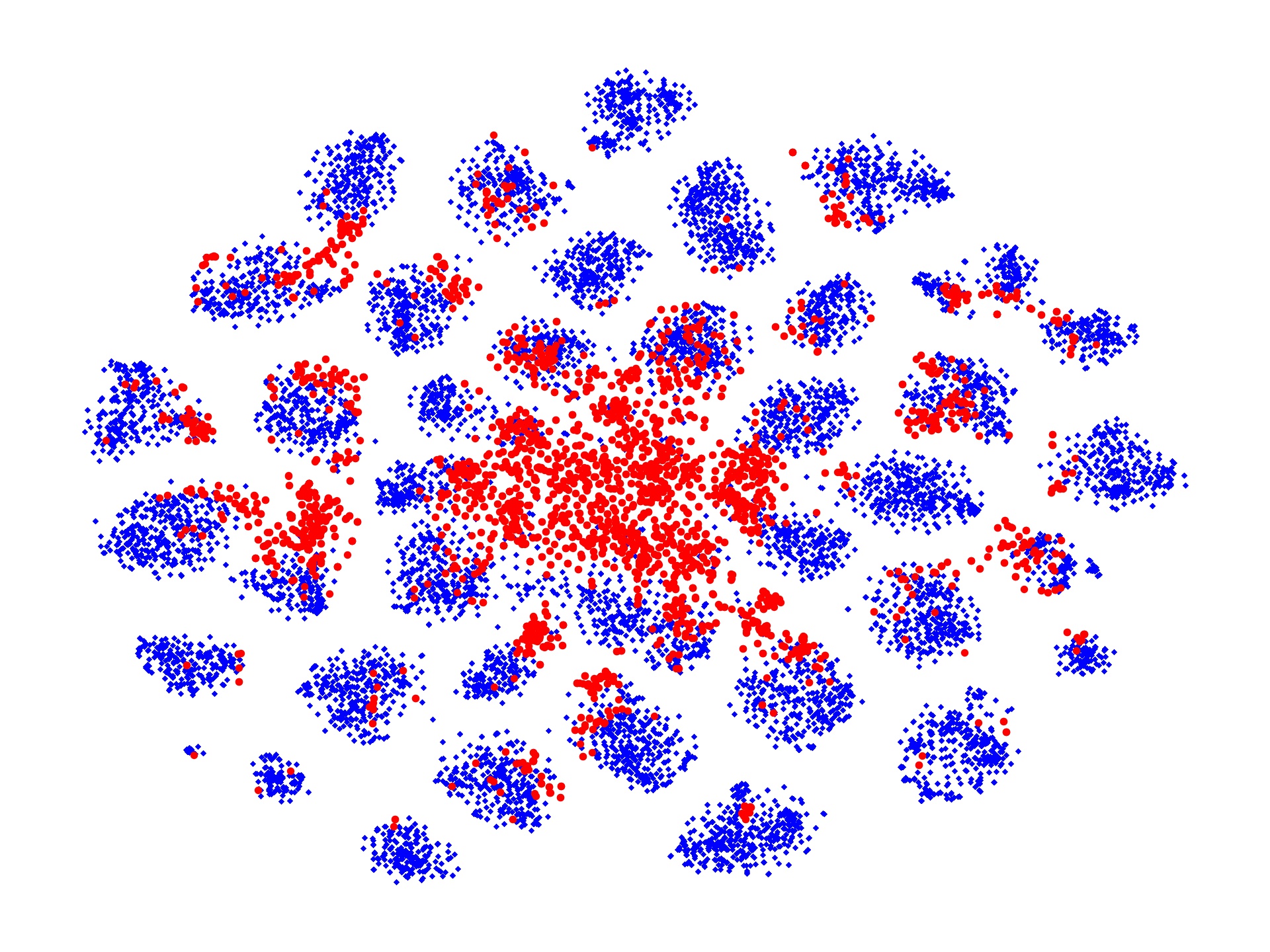}
        \caption{DAN}
    \end{subfigure}
      \hfill
    \begin{subfigure}{.18\linewidth}
        \centering
        \includegraphics[scale=0.16]{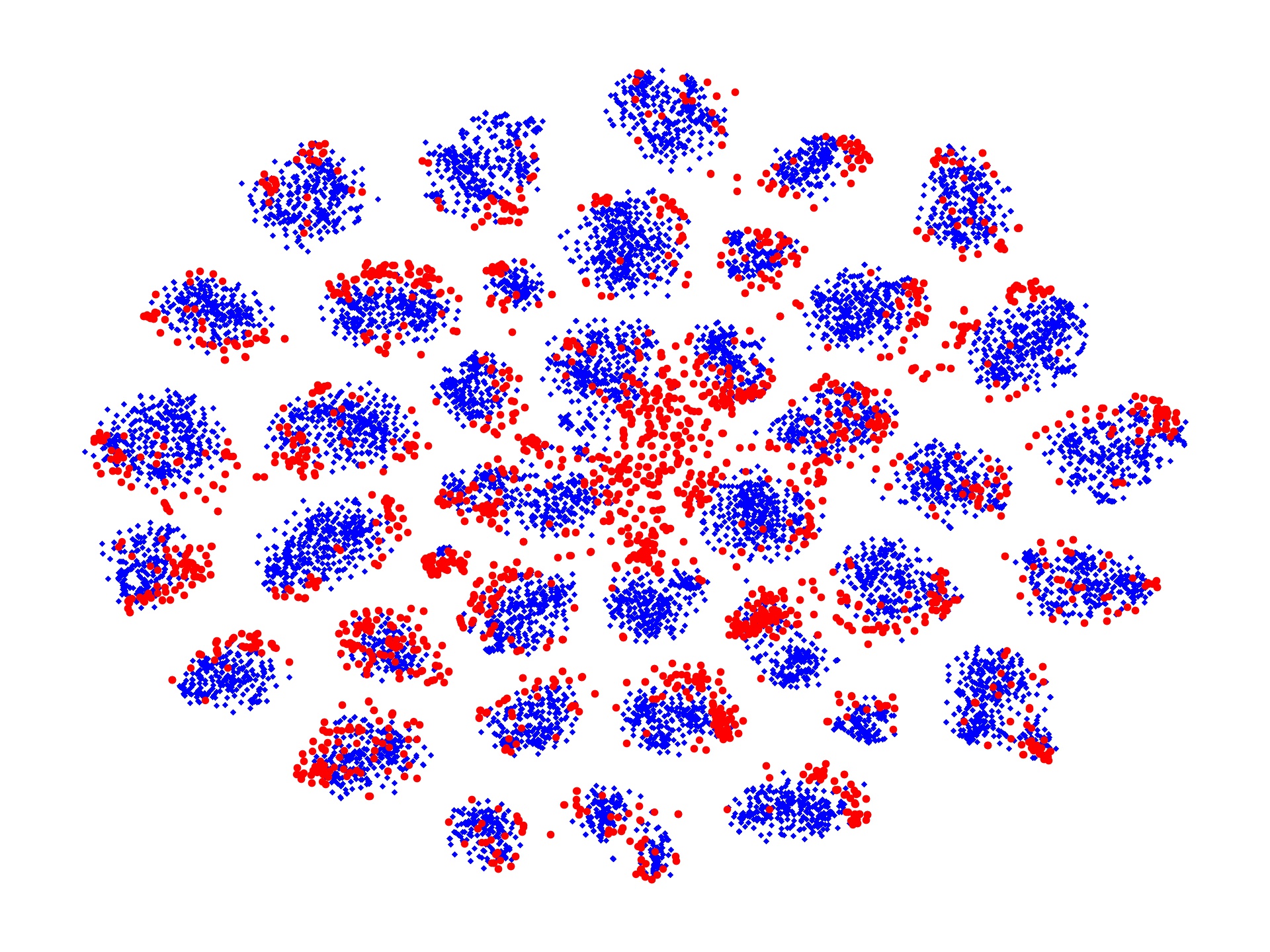}
        \caption{DANN}
    \end{subfigure}
        \hfill
    \begin{subfigure}{.18\linewidth}
        \centering
        \includegraphics[scale=0.16]{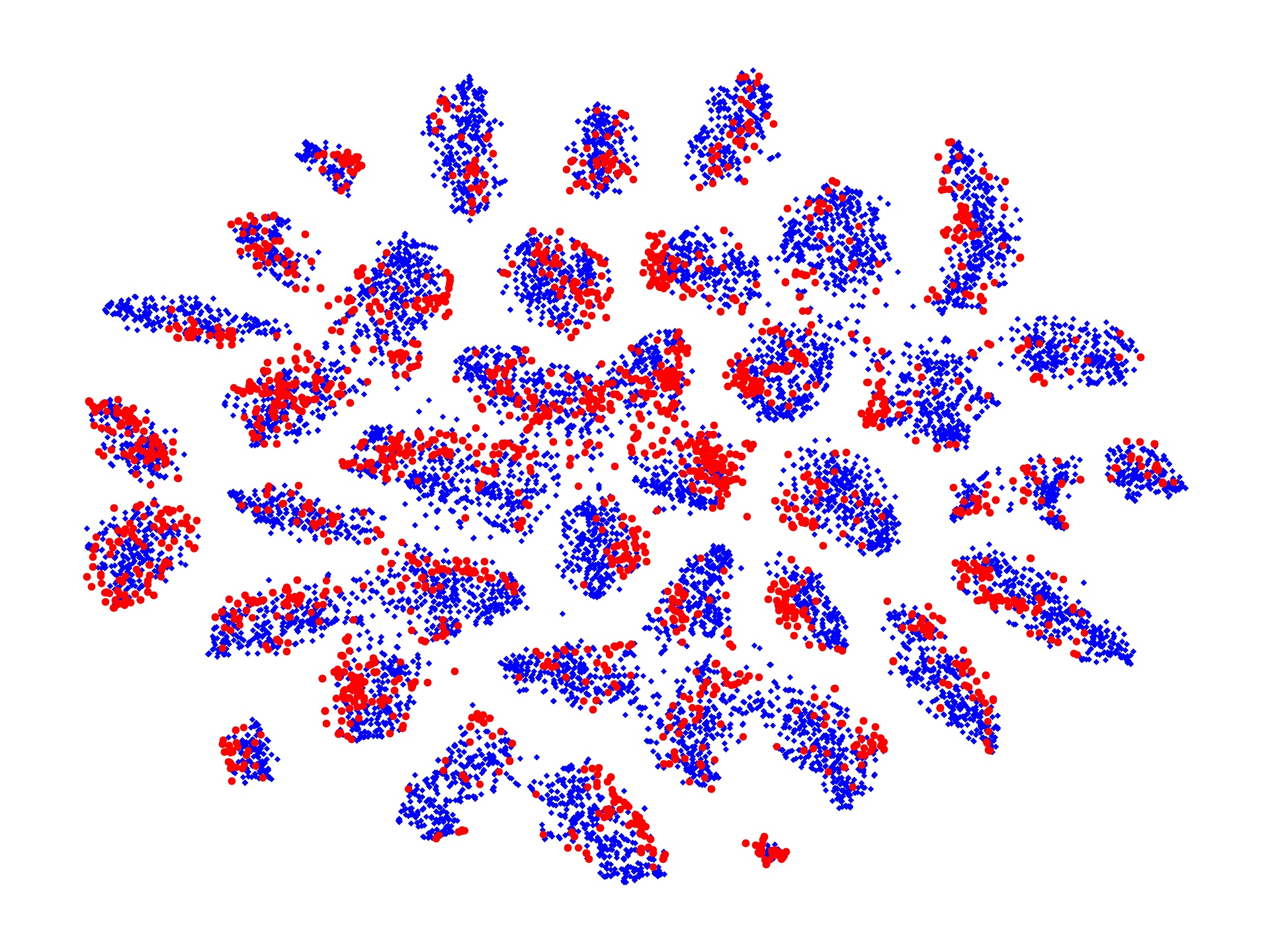}
        \caption{\textbf{\MODELNAME}}
    \end{subfigure}
    \caption{t-SNE visualization for features Source Only (baseline), DAN, DANN and \MODELNAME~on DomainNet task Real $\rightarrow$ Clipart. Blue and red points represents features from the source domain and target domain, respectively.
    \label{fig:tsne}}
\end{figure}

\begin{wrapfigure}{r}{0.50\textwidth}
    \centering
    \vspace{-0.1in}
    \includegraphics[scale=0.35]{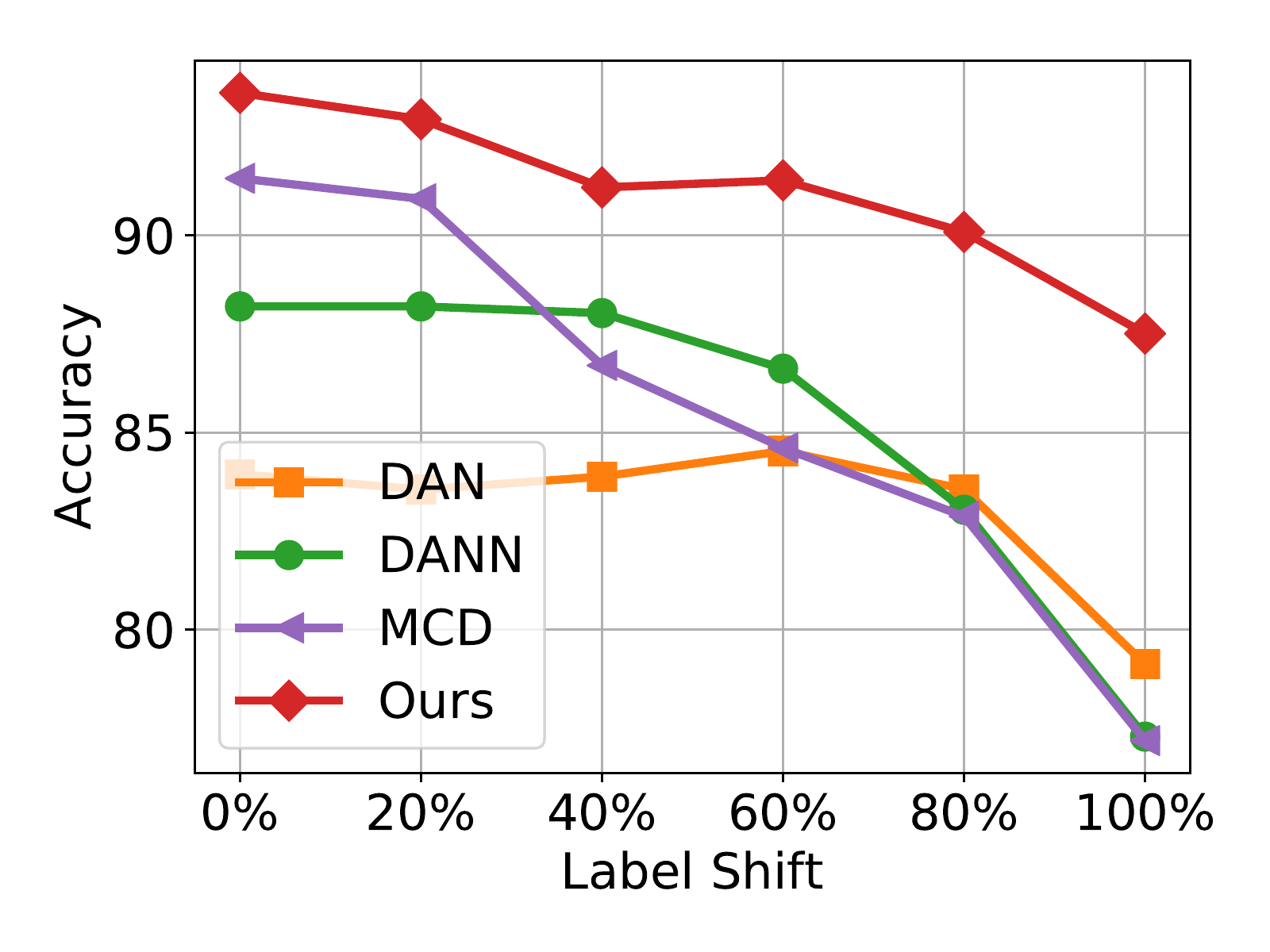}
    \vspace{-0.2in}
    \caption{\small Performance on USPS$\rightarrow$MNIST task with different degrees of \ssh{\YSHIFTNAME}. $0\%$ and $100\%$ denote the BS-BT and RS-UT settings respectively. Others are the linear interpolations of BS-BT and RS-UT.}
    \label{fig:label_shift}
    \vspace{-0.2in}
\end{wrapfigure}

\noindent \textbf{Different Degrees of \ssh{Label Shift}.}
In Section \ref{exp:total}, we only investigated the performance of each method under a certain degree of \ssh{\YSHIFTNAME} in each dataset. In this section, we investigate the effect of different degrees of \ssh{\YSHIFTNAME}. Specifically, we create 4 interval degrees of \ssh{\YSHIFTNAME} between the \textbf{BS-BT} (\textbf{B}lanced \textbf{S}ource and \textbf{B}lanced \textbf{T}arget) and RS-UT setting. To this end, we compute the proportions of each category by linear interpolation between its proportions in BS-BT and RS-UT. We denote BS-BT and RS-UT as $0\%$ and $100\%$ degree of \ssh{\YSHIFTNAME} respectively, and create datasets with \ssh{\YSHIFTNAME}s of $20\%, 40\%, 60\%$ and $80\%$ by linear interpolation. For fair comparison, all the datasets have the same total amount of samples. With these datasets, we evaluated the performance of different methods on the USPS $\rightarrow$ MNIST task. The results in Figure \ref{fig:label_shift} show that the performance of previous domain adaptation methods will be significantly affected by \ssh{\YSHIFTNAME}. For example, the accuracy of MCD drastically drops from \textbf{91.45}\% to \textbf{77.18}\%. In contrast, the performance of COAL is much more stable, which ranges between \textbf{93.42}\% and \textbf{88.12}\%. It shows that our framework is more robust to different degrees of \ssh{\YSHIFTNAME}.

\input{4.7_AblationStudy.tex}
\noindent \textbf{Feature Visualization.} In this section, we plot the learned features with t-SNE \cite{Maaten08visualizingdata} in Figure \ref{fig:tsne}. We investigate the Real to Clipart task in DomainNet experiment with ResNet-50 backbones. From (a)-(d), we observe that our method can better align source and target features in each category, while other methods either leave the feature distributions unaligned, or incorrectly aligned samples in different categories. These results further show the importance of \textit{prototype-based conditional feature alignment} for \ssh{\DANAME} task.

%% file: 4.2_DigitResult.tex
\begin{table}[t]
   \renewcommand\arraystretch{1.0}
  \scriptsize
  \centering 
  \begin{tabular}{lccccc} 
  \toprule[1.2pt]
    Methods  & \textbf{USPS$\rightarrow$MNIST}& \textbf{MNIST$\rightarrow$USPS}
          & \textbf{SVHN$\rightarrow$MNIST}& \textbf{SYN $\rightarrow$MNIST} & AVG  \\
    \midrule
    Source Only & 75.31$\pm$0.09 & 87.92$\pm$0.74 & 50.25$\pm$0.81 & \underline{85.74}$\pm$0.49 & 74.81  \\
    \midrule
    UAN (2019) & 55.72$\pm$2.06 & 83.23$\pm$0.75 & 50.20$\pm$1.75 & 71.26$\pm$2.82 & 65.10  \\
    ETN (2019) & 62.85$\pm$2.20 & 79.27$\pm$1.29 & 52.82$\pm$1.50 & 72.42$\pm$6.53 & 66.84  \\
    FDANN (2019) & 72.59$\pm$1.61 & 81.62$\pm$2.38 & 45.65$\pm$2.93 & 82.07$\pm$1.65 & 70.48  \\
    JAN (2017) & 75.75$\pm$0.75 & 78.82$\pm$0.93 & 53.21$\pm$3.94 & 75.64$\pm$1.42 & 70.86  \\
    BBSE (2018) & 75.01$\pm$3.68 & 78.84$\pm$10.73 & 49.01$\pm$2.02 & 85.69$\pm$0.71 & 72.14  \\
    BSP (2019) & 71.99$\pm$1.52 & 89.74$\pm$0.77 & 50.61$\pm$1.67 & 77.30$\pm$1.20 & 72.41  \\
    PADA (2018) & 73.66$\pm$0.15 & 78.59$\pm$0.23 & 54.13$\pm$1.61 & 85.06$\pm$0.60 & 72.86  \\
    MCD (2018) & 77.18$\pm$5.65 & 85.34$\pm$4.07 & 53.52$\pm$4.23 & 76.37$\pm$3.48 & 73.10  \\
    DAN (2015) & \underline{79.12}$\pm$1.34 & 87.15$\pm$1.71 & 53.63$\pm$1.80 & 80.89$\pm$2.00 & 75.20  \\
    DANN (2015) & 77.28$\pm$2.13 & \underline{91.88}$\pm$0.74 & \underline{57.16}$\pm$1.83 & 77.60$\pm$1.29 & \underline{75.98}  \\
    
    \midrule
    \textbf{\MODELNAME} (Ours) & \textbf{88.12$\pm$0.37} & \textbf{93.04$\pm$1.67} & \textbf{65.67$\pm$1.29} & \textbf{90.60$\pm$0.44} & \textbf{84.33} \\
    \bottomrule[1.2pt]
    \end{tabular}%
    \caption{Per-class mean accuracy on Digits. Our model achieves \textbf{84.33}\% average accuracy across four tasks, outperforming other evaluated methods.}
    \vspace{-0.2in}
    \label{tab:digit}%
\end{table}%

%% file: 4.3_OfficeHomeResult.tex
\begin{table}[t]
 \renewcommand\arraystretch{0.8}
  \scriptsize
  \centering
  \begin{tabular}{l*{7}{p{1.35cm}<{\centering}}} 
  \toprule[1.2pt]
    Methods  &\textbf{Rw}$\rightarrow$\textbf{Pr} & \textbf{Rw}$\rightarrow$\textbf{Cl} 
          & \textbf{Pr}$\rightarrow$\textbf{Rw} & \textbf{Pr}$\rightarrow$\textbf{Cl} & \textbf{Cl}$\rightarrow$\textbf{Rw} &\textbf{Cl}$\rightarrow$\textbf{Pr} &  AVG  \\          
    \midrule
    SourceOnly & 70.75  & 35.51  & 65.65  & 34.99  & 51.27  & 51.11  & 51.55  \\
    \midrule
    BSP (2019) & 66.15  & 23.48  & 65.42  & 20.81  & 34.54  & 31.04  & 40.24  \\
    PADA (2018) & 60.77  & 32.28  & 57.09  & 26.76  & 40.71  & 38.34  & 42.66  \\
    BBSE (2018) & 61.10  & 33.27  & 62.66  & 31.15  & 39.70  & 38.08  & 44.33  \\
    MCD (2018) & 66.18  & 32.32  & 62.66  & 28.40  & 41.41  & 38.59  & 44.93  \\
    DAN (2015) & 67.85  & 38.17  & 66.86  & 34.24  & 52.95  & 51.64  & 45.02  \\
    UAN (2019) & 70.85  & 41.15  & 67.26  & 36.82  & 56.24  & 55.77  & 48.62  \\
    ETN (2019) & 71.69  & 34.03  & \underline{70.45}  & \textbf{40.74}  & \textbf{60.48}  & 55.19  & 52.14  \\
    FDANN (2019) & 68.56  & 40.57  & 67.32  & 37.33  & 55.84  & 53.67  & 53.88  \\
    JAN (2017) & 71.22  & \underline{43.12}  & 68.20  & 37.03  & 57.97  & 56.80  & 55.72  \\
    DANN (2015) & 71.78  & \textbf{46.08}  & 67.98  & 39.45  & 58.40  & \underline{57.39}  & \underline{56.85}  \\
    \midrule
    \textbf{\MODELNAME} (Ours) & \textbf{73.65} & 42.58 & \textbf{74.46} & \underline{40.61} & \underline{59.22} & \textbf{62.71} & \textbf{58.87} \\
    \bottomrule[1.2pt]
    \end{tabular}%
    \caption{Per-class mean accuracy on Office-Home dataset. Our model achieve \textbf{58.87}\% average accuracy across six tasks. }
  \vspace{-0.4in}
  \label{tab:OH}%
\end{table}%

%% file: 4.5_DomainNetResult.tex
\begin{table}[t]
 \renewcommand\arraystretch{0.8}
  \scriptsize
  \centering
  \begin{tabular}{l*{13}{p{0.75cm}<{\centering}}} 
  \toprule[1.2pt]
    Method & \textbf{R}$\rightarrow$\textbf{C}& \textbf{R}$\rightarrow$\textbf{P} & \textbf{R}$\rightarrow$\textbf{S} & \textbf{C}$\rightarrow$\textbf{R} & \textbf{C}$\rightarrow$\textbf{P} & \textbf{C}$\rightarrow$\textbf{S} &\textbf{P}$\rightarrow$\textbf{R} & \textbf{P}$\rightarrow$\textbf{C} & \textbf{P}$\rightarrow$\textbf{S} & \textbf{S}$\rightarrow$\textbf{R}& \textbf{S}$\rightarrow$\textbf{C} & \textbf{S}$\rightarrow$\textbf{P} & AVG  \\
    \midrule
    Baseline & 58.84  & 67.89  & 53.08  & 76.70  & 53.55  & 53.06  & 84.39  & 55.55  & 60.19  & 74.62  & 54.60  & 57.78  & 62.52  \\
    \midrule
    BBSE & 55.38  & 63.62  & 47.44  & 64.58  & 42.18  & 42.36  & 81.55  & 49.04  & 54.10  & 68.54  & 48.19  & 46.07  & 55.25  \\
    PADA & 65.91  & 67.13  & 58.43  & 74.69  & 53.09  & 52.86  & 79.84  & 59.33  & 57.87  & 76.52  & 66.97  & 61.08  & 64.48  \\
    MCD & 61.97  & 69.33  & 56.26  & 79.78  & 56.61  & 53.66  & 83.38  & 58.31  & 60.98  & 81.74  & 56.27  & 66.78  & 65.42  \\
    DAN & 64.36  & 70.65  & 58.44  & 79.44  & 56.78  & 60.05  & 84.56  & 61.62  & 62.21  & 79.69  & 65.01  & 62.04  & 67.07  \\
    FDANN & 66.15  & 71.80  & 61.53  & 81.85  & 60.06  & 61.22  & 84.46  & 66.81  & 62.84  & 81.38  & 69.62  & 66.50  & 69.52  \\
    UAN & 71.10  & 68.90  & 67.10  & 83.15  & 63.30  & 64.66  & 83.95  & 65.35  & 67.06  & 82.22  & 70.64  & 68.09  & 72.05  \\
    JAN & 65.57  & \underline{73.58}  & 67.61  & 85.02  & 64.96  & 67.17  & \underline{87.06}  & 67.92  & 66.10  & 84.54  & 72.77  & 67.51  & 72.48  \\
    ETN & 69.22  & 72.14  & 63.63  & \underline{86.54}  & 65.33  & 63.34  & 85.04  & 65.69  & 68.78  & 84.93  & 72.17  & 68.99  & 73.99  \\
    BSP & \underline{67.29}  & 73.47  & 69.31  & 86.50 & \underline{67.52}  & \underline{70.90}  & 86.83  & 70.33  & 68.75  & 84.34  & 72.40  & \textbf{71.47}  & 74.09  \\
    DANN & 63.37  & 73.56  & \textbf{72.63} & 86.47  & 65.73  & 70.58  & 86.94 & \underline{73.19} & \underline{70.15} & \underline{85.73} & \textbf{75.16} & 70.04  & \underline{74.46}  \\
    \midrule
    \textbf{Ours} & \textbf{73.85} & \textbf{75.37} & \underline{70.50}  & \textbf{89.63}  & \textbf{69.98} & \textbf{71.29} & \textbf{89.81} & 68.01  & \textbf{70.49}  & \textbf{87.97}  & \underline{73.21}  & \underline{70.53} & \textbf{75.89} \\
    \bottomrule[1.2pt]
    \end{tabular}%
    \caption{Per-class mean accuracy on DomainNet dataset with natural label shifts. Our method achieve \textbf{75.89}\% average accuracy across the 12 experiments. Note that DomainNet contains about 0.6 million images, it is non-trivial to have even one percent performance boost.  }
    \vspace{-0.4in}
  \label{tab:DM}%
\end{table}%

%% file: 4.6_BalancedSamplingResult.tex
\begin{table}[t]
 \renewcommand\arraystretch{1.0}
 \scriptsize
  \centering
  \begin{tabular}{l*{12}{p{0.97cm}<{\centering}}} 
  \toprule[1.2pt]
   \multirow{2}{*}{Methods}  & \multicolumn{2}{c}{U$\rightarrow$M}& \multicolumn{2}{c}{S$\rightarrow$M}
          & \multicolumn{2}{c}{Pr$\rightarrow$Cl} & \multicolumn{2}{c}{Cl$\rightarrow$Rw} & \multicolumn{2}{c}{R$\rightarrow$S}   \\
          & {w/o} & with & w/o & with  
          & w/o & with & w/o & with
          & w/o & with 
          \\          
    \midrule
    SourceOnly & 71.35 & \textbf{75.31} & 50.35 & 50.25 & 34.99 & \textbf{34.99} & 50.64 & \textbf{51.11} & 50.16 & \textbf{53.08} \\
    DAN & 64.81  & \textbf{79.12}  & 22.05  & \textbf{53.63}  & 32.93 & \textbf{34.24} & 45.18  & \textbf{51.64}  & 64.78  & 58.44  \\
    DANN & 42.77  & \textbf{77.28}  & 27.60  & \textbf{57.16}  & 35.17 & \textbf{39.45} & 47.19  & \textbf{58.40}  & 68.92  & \textbf{72.63} \\
    MCD & 20.15  & \textbf{77.18}  & 44.83  & \textbf{53.52}  & 33.06 & 28.40 & 49.57  & 41.41  & 58.50  & 56.26  \\
    \midrule
    \textbf{\MODELNAME} & 87.50  & \textbf{88.12}  & 60.12  & \textbf{65.67}  & 34.03 & \textbf{40.61}  & 57.67  & \textbf{59.22}  & 59.23  & \textbf{70.50}  \\
    \bottomrule[1.2pt]
    \end{tabular}%
    \caption{The performance of five models \textit{w.} or \textit{w/o.} source balanced sampler. We observe a significant performance boost when the source balanced sampler is applied, both for our model and the compared baselines, demonstrating the effectiveness of source balanced sampler to \DANAME~task.}
  \vspace{-0.4in}
  \label{tab:balanced}%
\end{table}%

%% file: 4.7_AblationStudy.tex
\begin{table}[t]
 \renewcommand\arraystretch{1.0}
  \scriptsize
  \centering
  \begin{tabular}{l*{9}{p{1.1cm}<{\centering}}} 
  \toprule[1.2pt]
    Methods  & U$\rightarrow$M & M$\rightarrow$U
          &S$\rightarrow$M & Cl$\rightarrow$Rw & Pr$\rightarrow$Rw & R$\rightarrow$C & R$\rightarrow$P & P$\rightarrow$R & AVG \\          
    \midrule
    w/o $\mathcal{L}_{ST}$  & 85.22 & 85.94 & 55.17 & 58.38 & 69.39 & 71.92 & 74.39 & 77.45 &  72.23\\
    w/o $\mathcal{L}_{H}$  & 85.57 & 92.28 & 63.34 & 58.17 & 72.11 & 71.34 & 69.92 & 87.14 & 74.98 \\
    \midrule
    \textbf{\MODELNAME} & 88.12 & 93.04 & 65.67 & 59.22 & 74.46 & 73.85 & 75.37 & 89.81 & 77.44 \\
    \bottomrule[1.2pt]
    \end{tabular}%
    \caption{Ablation study of different objectives in our method. We randomly select 8 sets of experiments to perform the ablation study. 
    }
  \label{tab:ablation}%
\end{table}%

%% file: 5_conclusion.tex
\section{Conclusion}

In this paper, we first propose the  \ssh{\DANAMEFULL (\textbf{\DANAME})} setting and demonstrate its importance in practical scenarios. \ssh{Then we provide the first benchmark of this problem, and conduct a comprehensive empirical analysis on recent domain adaptation methods. The result shows that most existing methods are fragile in the face of \DANAME, which prevents them from being practically applied. Based on theoretical motivations, we propose a feature distribution and label distribution co-alignment framework, which empirically works well as a baseline for future research.}

\ssh{We believe this work takes an important step towards applicable domain adaptation. We hope the provided benchmarks, empirical results and baseline model would stimulate and facilitate future works to design robust algorithms that can handle more realistic problems. An interesting research direction would be better detecting and correcting label shift under feature shift.}

%% file: 6_appendix.tex
\setcounter{section}{0}
\setcounter{equation}{0}
\setcounter{figure}{0}
\setcounter{table}{0}
\setcounter{page}{1}

\section{Creation Details for Label Shift}
In order to create unbalanced label distribution in each dataset, inspired by \cite{openlongtailrecognition}, we follow the Paredo distribution set different proportions for each category. By using this distribution, we can create \textit{long-tailed} label distribution, which is frequently seen in real applications and benchmarks \cite{cvpr18_inat,DomainNet}.

The shape of Paredo distribution \cite{Pareto} is controlled by parameter $\alpha$. Because different datasets have different amount of samples, to avoid making some classes in the unbalanced dataset to have too few samples, we use a different parameter for each dataset. Specifically, we set $\alpha=1$ for Digits dataset, and $\alpha=100$ for Office-Home dataset. 

We further assign each computed proportion to each category by following the descending order of the original class index. Specifically, in the target domain of RS-UT, we assign the $k_{th}$ largest propotion to class $k-1$, with class index starting from 0. For Digits, we set the index of each class directly as the digit it represents. For Office-Home, we set the class index in alphabetical ascending order.

Please refer to the attached code for more details about the detailed list of image files for each of the domain in the three datasets.

\section{COAL Implementation Details}
For the Digits dataset, we adopt the network architecture proposed by \cite{MCD_2018}. We adopt SGD with the momentum of 0.9 and learning rate of 0.01 for the linear classifier and 0.001 for all other layers. The batch size is set as 32 for samples from each domain. For USPS$\rightarrow$MNIST, MNIST$\rightarrow$USPS and SynD$\rightarrow$MNIST, we set the $\alpha=0.1$ in Equation 6 and set $k_0=20, k_{step}=5, k_{max}=50$. For the more challenging SVHN$\rightarrow$MNIST, we set $k_0=5, k_{step}=5, k_{max}=10$.

For OfficeHome and DomainNet, we utilize ResNet-50 \cite{He2015DeepRL} as our backbone network, and replace the last fully-connected layer with a randomly initialized N-way classifier layer (for N categories). We also use SGD with momentum of 0.9 while setting the learning rate to be 0.001 for linear layers and 0.0001 for all the other layers. The batch size is set as 16 for each domain. We set the $\alpha=0.1$ in Equation 6 and set $k_0=5, k_{step}=5, k_{max}=30$ as the parameters for self-training selection policy. 

\section{Hyper-parameter setting for Compared Methods}
We tune the hyper-parameters of each method on Painting $\rightarrow$ Clipart task in DomainNet. Specifically, for DAN, JAN, FDANN and DANN we tune the weight $\alpha$ of the marginal feature alignment loss. We empirically find that these method achieve better performance when we set $\alpha$ to be 5-10 times \textit{lower} than default. Intuitively, it means that we can achieve better performance under generalized domain shift setting if we relax the strength of marginal feature alignment. For MCD we tune the number of feature generator updating times $n$.

\section{Detailed information for datasets}
We provide detailed information for datasets in Table \ref{tab:information}.

\input{6.4_dataset_information.tex}

%% file: 6.4_dataset_information.tex
\begin{table}[htbp]
 \renewcommand\arraystretch{1.0}
  \centering 
  \begin{tabular}{l*{5}{p{2.0cm}<{\centering}}} 
  \toprule[1.2pt]
    \multicolumn{6}{c}{Digits} \\
    \midrule
    Splits  & USPS & MNIST & SVHN & SYN & Total \\
    \midrule
    Train &  12,144 & 1,550 & 10,395 & 107,005 & 118,950 \\
    Test &  2,181 & 459 & 3,554 & 2,114 & 8,308 \\
    \midrule
    \multicolumn{6}{c}{Office-Home} \\
    \midrule
    Splits  & Real World & Product & Clipart && Total \\
    \midrule
    Total & 1,253 & 2,045 & 1,017 &  & 4,315 \\
    \midrule
    \multicolumn{6}{c}{DomainNet} \\
    \midrule
    Splits  & Real & Painting & Clipart & Sketch & Total \\
    \midrule
    Train &  16,141 & 6,727 & 3,707 & 5,537 & 32,112 \\
    Test &  6,943 & 2,909 & 1,616 & 2,399 & 13,867 \\
    
    \bottomrule[1.2pt]
    \end{tabular}%
    \caption{Detailed information for datasets}
    \label{tab:information}%
\end{table}%